\documentclass[]{fairmeta}

\usepackage{amsfonts}
\usepackage{nicefrac}
\usepackage{pifont}
\usepackage{amsmath}
\usepackage{amssymb}
\usepackage{amsthm}
\usepackage{mathtools}
\usepackage{xspace}
\usepackage{enumitem}
\usepackage{makecell}
\usepackage{threeparttable}
\usepackage[textsize=tiny]{todonotes}

\usepackage{xcolor}
\usepackage{pifont}

\definecolor{cmarkgreen}{HTML}{006400} 
\definecolor{xmarkred}{HTML}{8B0000}   

\newcommand{\cmark}{\textcolor{cmarkgreen}{\ding{51}}}
\newcommand{\xmark}{\textcolor{xmarkred}{\ding{55}}}

\newcommand{\R}{\mathbb{R}}

\newcommand{\KL}{\mathrm{KL}}

\DeclareMathOperator*{\argmin}{arg\,min}

\newcommand{\membuilder}{MemBuilder\xspace}
\newcommand{\memt}{Mem-T\xspace}
\newcommand{\memalpha}{Mem-$\alpha$\xspace}

\def\gH{{\mathcal{H}}}

\def\gM{{\mathcal{M}}}

\def \policy{{\pi_\theta}}

\newcommand{\mypar}[1]{\noindent \textbf{#1}}

\newcommand{\ourshort}{AttriMem}

\crefformat{section}{\S#2#1#3}
\crefformat{subsection}{\S#2#1#3}
\crefformat{subsubsection}{\S#2#1#3}
\crefformat{paragraph}{\P#2#1#3}
\crefformat{subparagraph}{\P#2#1#3}
\crefmultiformat{section}{\S#2#1#3}{ and~\S#2#1#3}{, \S#2#1#3}{, and~\S#2#1#3}
\crefmultiformat{subsection}{\S#2#1#3}{ and~\S#2#1#3}{, \S#2#1#3}{, and~\S#2#1#3}
\crefmultiformat{subsubsection}{\S#2#1#3}{ and~\S#2#1#3}{, \S#2#1#3}{, and~\S#2#1#3}
\crefmultiformat{paragraph}{\P\P#2#1#3}{ and~#2#1#3}{, #2#1#3}{, and~#2#1#3}
\crefmultiformat{subparagraph}{\P\P#2#1#3}{ and~#2#1#3}{, #2#1#3}{, and~#2#1#3}
\crefrangeformat{section}{\mbox{\S\S#3#1#4--#5#2#6}}
\crefrangeformat{subsection}{\mbox{\S\S#3#1#4--#5#2#6}}
\crefrangeformat{subsubsection}{\mbox{\S\S#3#1#4--#5#2#6}}
\crefrangeformat{paragraph}{\mbox{\P\P#3#1#4--#5#2#6}}
\crefrangeformat{subparagraph}{\mbox{\P\P#3#1#4--#5#2#6}}
\crefname{part}{Part}{Parts}
\Crefname{part}{Part}{Parts}
\crefname{chapter}{Ch.}{Ch.}
\Crefname{chapter}{Ch.}{Ch.}
\crefname{footnote}{Fn.}{Fn.}
\Crefname{footnote}{Fn.}{Fn.}
\crefname{figure}{Figure}{Figures}
\crefname{table}{Table}{Tables}
\crefname{subfigure}{Figure}{Figures}
\Crefname{subfigure}{Figure}{Figures}
\crefname{appsec}{Appendix}{Appendices}
\Crefname{appsec}{Appendix}{Appendices}
\crefname{algocf}{Algorithm}{Algorithms}
\Crefname{algocf}{Algorithm}{Algorithms}

\crefname{appendix}{Section}{Sections}
\crefalias{subappendix}{Section}

\definecolor{darkpink}{rgb}{0.8, 0.125, 0.5}
\definecolor{darkblue}{rgb}{0, 0, 0.5}
\definecolor{brandblue}{rgb}{0.34, 0.7, 1}

\tcbset {
  base/.style={
    arc=0mm,
    bottomtitle=0.5mm,
    boxrule=0mm,
    colbacktitle=black!10!white,
    coltitle=black,
    fonttitle=\bfseries,
    left=2.5mm,
    leftrule=1mm,
    right=3.5mm,
    title={#1},
    toptitle=0.75mm,
    width=.95\textwidth
  }
}

\newtcolorbox{mainbox}[1]{
  colframe=darkpink,
  base={#1},
  width=\linewidth,
}

\setuniversityname{}

\title{AttriMem: Attribution-Guided Process Feedback for Agent Memory Construction}

\author[1]{Qinfeng Li\textsuperscript{$\dagger$}}
\author[1]{Yuntai Bao\textsuperscript{$\dagger$}}
\author[1]{Xinyan Yu\textsuperscript{$\dagger$}}
\author[1]{Hongze Chen}
\author[1]{Yanming Liu}
\author[2]{Huifeng Zhu}
\author[3]{Yier Jin}
\author[1]{Jintao Chen}
\author[1]{Wenqi Zhang}
\author[1]{Xuhong Zhang\textsuperscript{$*$}}

\affiliation[1]{Zhejiang University}
\affiliation[2]{Washington University in St. Louis}

\affiliation[3]{University of Science and Technology of China}

\abstract{

Effective memory is crucial for LLM agents, yet how to construct it remains a challenge. A memory-construction policy decides what information to extract, store, update, compress, or discard as interactions accumulate. Heuristic memory methods rely on subjective, task-specific rules, which can misalign with downstream objectives and limit cross-task adaptability. RL-based methods, by contrast, learn from task feedback but mainly use outcome- or action-level rewards. These coarse signals indicate task success but cannot identify which intermediate memory contents support the final answer, creating a \textit{fine-grained credit-assignment bottleneck}. However, constructing such process feedback is prohibitively difficult because intermediate memory decisions lack unique ground-truth targets, while the appropriate credit varies with the agent's uncertain reasoning trajectory and therefore cannot be specified in advance.
We propose \textbf{\ourshort{}}, an attribution-guided process-feedback framework for learning memory-construction policies with RL. \ourshort{} augments the global outcome reward with local rewards derived from token-level contributions to the final answer.
Experiments on long-horizon dialogue question answering show that \ourshort{} outperforms retrieval-based, heuristic, and RL-based baselines, generalizes across benchmarks and answer models, and stabilizes RL optimization.

}

\begin{document}
\thispagestyle{firstheader}
\maketitle
\pagestyle{empty}
\begingroup
\renewcommand{\thefootnote}{\fnsymbol{footnote}}
\footnotetext[2]{These authors contributed equally.}
\footnotetext[1]{Corresponding author: \texttt{zhangxuhong@zju.edu.cn}.}
\endgroup

\section{Introduction}
\label{sec:intro}
As LLM agents are used in tasks with long interactions and large amounts of information, they need to remember and use past interactions to make good decisions~\citep{park2023generative,zhong2024memorybank,wu2024longmemeval}. This is challenging because the information needed to answer questions is often scattered across sessions and mixed with irrelevant details~\citep{maharana2024locomo,liu2024lost}. To handle such long interaction histories, a memory-construction policy incrementally builds an external memory store by deciding what information to extract, store, update, merge, compress, or discard. 
Therefore, \textbf{\textit{a key problem is how to learn such a memory-construction policy (i.e., memory learning) to preserve information useful for future tasks while filtering out unnecessary context.}}

However, as shown in Table~\ref{tab:paradigm_comparison}, existing approaches to constructing such memory remain limited in several important aspects. For example, \textbf{heuristic memory methods}~\citep{gam2024,amem,lightmem2026,mem02024,lewis2020rag} rely on manually designed rules to determine what information should be stored, retained, or retrieved, making memory construction highly subjective. On one hand, such subjectivity can easily lead to objective mismatch~\citep{zhang2026adaptive}: information that appears salient under the designer's intuition may not actually matter for downstream decisions. On the other hand, these methods are often tightly coupled to a particular task setting, which limits their generalizability~\citep{zhou2026we}. When the task objective changes, the original memory policy often fails to adapt flexibly, requiring further manual adjustment.

\newcommand{\pmark}{\ensuremath{\triangle}} 

\begin{table*}[t]
\centering
\small
\setlength{\tabcolsep}{8pt}
\renewcommand{\arraystretch}{1}
\resizebox{\textwidth}{!}{
\begin{tabular}{lcccc}
\toprule
Paradigm (exemplar)
& Learned from Task Feedback
& Process Feedback
& Additional Reward Signal
& \textbf{Token-Level Feedback} \\
\midrule
Heuristic Memory~\citep{zhong2024memorybank}
& \xmark & \xmark & \xmark & \xmark \\

Task-Performance RL~\citep{memalpha2024}
& \cmark & \xmark & \xmark & \xmark \\

Outcome-Redistributed Rewards~\citep{memt2026}
& \cmark & \cmark & \xmark & \xmark \\

QA-Derived Action Rewards~\citep{membuilder2026}
& \cmark & \cmark & \cmark & \xmark \\

\ourshort{} (ours)
& \cmark & \cmark & \cmark & \cmark \\
\bottomrule
\end{tabular}
}
\caption{
Comparison of memory-learning paradigms. 
Among the compared paradigms, \ourshort{} is the only paradigm that learns from task feedback, introduces feedback beyond outcomes, and provides action- and token-level guidance for memory learning.
}
\label{tab:paradigm_comparison}
\end{table*}

\textbf{RL-based memory-learning methods}~\citep{memt2026,membuilder2026,memalpha2024}, by contrast, learn memory behavior from task-grounded feedback, enabling optimization of memory policies for diverse tasks. However, existing methods still largely rely on final task performance for supervision, leaving intermediate memory actions with sparse rewards and creating a fine-grained credit-assignment bottleneck that hinders effective policy learning: such signals only indicate whether the task succeeds but provide little guidance on which memory actions are responsible for success or failure. A representative paradigm is \textbf{task-performance RL}, which optimizes memory construction using downstream performance signals, such as answer correctness~\citep{memalpha2024}. Although such rewards make memory learning task-grounded, they offer little direct guidance for intermediate memory actions, because an answer-only signal cannot specify how each memory action contributed to that outcome. To provide feedback for intermediate processes, \textbf{outcome-redistributed reward methods}, such as Mem-T~\citep{memt2026}, redistribute final feedback to intermediate nodes through sampled memory-action trees, thereby propagating the outcome rewards of leaf trajectories to their ancestor memory-action nodes. However, this feedback fundamentally remains a redistribution of the final feedback and does not introduce any additional information, leaving the feedback still sparse.


More advanced \textbf{QA-derived action-reward methods}, such as MemBuilder~\citep{membuilder2026}, synthesize auxiliary QA pairs and reward memory actions according to whether their contents are retrieved and help answer these questions correctly, thereby providing richer reward signals for memory learning.
However, these rewards are still aggregated from final QA performance, rather than derived from direct evidence of how each intermediate memory action contributes to the answer, making the supervision insufficiently fine-grained.
In particular, an action-level signal evaluates a memory action only as a whole and therefore provides no targeted guidance for improving its internal content, even though its utility may depend on only a few critical tokens or spans, such as an entity, value, or timestamp.
This motivates finer-grained feedback that can evaluate and improve the specific content produced by each action, down to individual tokens.

These limitations expose a fundamental challenge in existing RL-based memory learning. Memory actions ideally require direct and fine-grained feedback, yet such feedback is intrinsically difficult to obtain because memory construction is itself a long-horizon problem over long contexts. The \textbf{central difficulty} is to determine which pieces of evidence actually support the ongoing reasoning trajectory and the final answer. First, existing memory benchmarks rarely provide ground-truth process signals for intermediate memory decisions. More fundamentally, \textbf{\textit{constructing such ground-truth is prohibitively difficult}}, because intermediate memory decisions do not have a unique correct target: which memory evidence should receive credit depends on the reasoning trajectory taken by the agent, and this trajectory is highly uncertain.
Consequently, \textbf{\textit{no universal and static process ground-truth can faithfully supervise intermediate memory.}}

This motivates us to seek an alternative source of process supervision: can task performance be converted into dense, fine-grained token-level credit signals for memory-process outputs? \textbf{Our key insight is to use the final answer to find the useful tokens in the intermediate memory process and convert their contributions into process rewards.}

Building on this insight, we propose \ourshort{}, an attribution-guided process-feedback framework for RL-based memory learning. During training, \ourshort{} follows the standard memory learning pipeline: the policy constructs intermediate memory outputs, a fixed retrieval-and-answering interface produces the final answer, and an outcome reward evaluates overall task performance. Beyond this global reward, \ourshort{} quantifies how much each token in the intermediate memory outputs contributes to the final answer. Concretely, it masks different subsets of memory tokens and uses the resulting changes in the final-answer score to estimate each token's contribution. These token-level scores are then mapped back to the memory process that produced them and converted into local process rewards. As a result, the policy receives feedback not only on overall task success, but also on the specific content within each memory action.

The \textbf{contributions} of our work are summarized as follows.
\begin{enumerate}[nosep, leftmargin=*]
\item We identify a fine-grained credit-assignment bottleneck in memory learning: existing RL-based methods provide only outcome- or action-level feedback, leaving the utility of specific memory contents under-specified.

\item We analyze why fine-grained process supervision is difficult to obtain: memory benchmarks usually lack ground-truth process signals for intermediate memory, and such ground-truth is prohibitively difficult to construct.


\item We propose \ourshort{}, an attribution-guided process-feedback framework that treats the final answer as a credit-assignment target and traces support for the final answer back to intermediate memory process rewards.

\item Extensive experiments show that token-level attribution enables more effective memory-policy optimization than outcome-only or action-level rewards, allowing \ourshort{} to achieve stronger task performance and more reliable memory-construction behavior.

\end{enumerate}




\section{Preliminaries}

\subsection{Problem Setting and Scope}
In this paper, we address the task of \textit{long-term dialogue question answering}.
Given a conversation history consisting of $T$ sessions $\mathcal{H} = \{ h_t \}_{t=1}^{T}$, where each session $h_t$ comprises multi-turn utterances at timestamp $t$, and a question $q$ posed after timestamp $T$, the goal is to generate an accurate answer based on information distributed across the entire history.

To isolate memory-construction quality from answer-model adaptation and support reusable, persistent memory, we adopt a modular architecture with a \textit{memory agent} and an \textit{answer model} used in recent memory-learning systems~\citep{membuilder2026}.
This separation reflects their different functional roles: memory construction is persistent and incrementally performed as interactions accumulate, whereas answer generation is query-specific.
Accordingly, \textbf{our goal} is to improve the memory agent rather than the answer model, while treating memory retrieval and answer generation as fixed components of the learning environment.
This modular design allows the same memory to serve future questions and different answer models, and enables a compact specialized policy to maintain memory while reserving a stronger model for reasoning.

The system therefore operates in two phases:

\begin{itemize}[nosep, leftmargin=2em]
    \item \textbf{Memory construction (memory agent).}
    An external memory bank $\mathcal{M}$ compresses and organizes historical information for selective retrieval.
    At each session $t$, a learnable memory-construction policy updates $\gM_t$ by choosing memory actions $a_t^{\mathrm{mem}}$ based on session $h_t$ and previous memory $\gM_{t-1}$.
    We denote by $\Delta \gM_t$ the textual memory content produced or modified by this update.
    Importantly, the policy does not observe the future question $q$, but incrementally constructs persistent memory from the dialogue history.

    \item \textbf{Question answering with memory retrieval (answer model).}
    To answer question $q$, a fixed retriever retrieves relevant records from $\gM_T$ and an answer model (e.g., Claude) generates an answer $\hat{y}$.
    The two components interact only through the external memory bank: the memory-construction policy organizes historical information, while the answer model performs query-specific reasoning over the retrieved records.
\end{itemize}

Accordingly, our scope is memory-construction behavior within long-term dialogue question answering, rather than a universally task-independent memory policy.

We optimize the memory-construction trajectory:
\begin{equation}
    \tau^{\mathrm{mem}} \coloneqq \{(h_t, \gM_{t-1}, a_t^{\mathrm{mem}}, \Delta \gM_t)\}_{t=1}^{T}.
\end{equation}
The \textbf{core challenge} is that rewards for intermediate memory actions are delayed (they arrive after fixed retrieval and answering), and final task success/failure provides no direct signal about which memory action was beneficial or detrimental.

\subsection{Memory Architecture}
\label{subsec:memory-architecture}

Here we provide a general description of the generic memory architecture for question answering in the context of multi-turn, long-horizon conversations, since the designs of recent memory agents are homogeneous at the high-level~\citep{membuilder2026,memt2026,zhou2026mem}.

Following the design of \citet{membuilder2026}, our memory architecture consists of four specialized memory modules:
$\mathcal{M}^{\mathrm{core}}$: Core memory, fixed-size user profile (identity, preferences, relationships);
$\mathcal{M}^{\mathrm{epi}}$: Episodic memory, time-stamped event records with temporal chaining;
$\mathcal{M}^{\mathrm{sem}}$: Semantic memory, factual knowledge about user-specific entities;
$\mathcal{M}^{\mathrm{proc}}$: Procedural memory, step-by-step processes and workflows.

Each component $m \in \{\mathrm{core}, \mathrm{epi}, \mathrm{sem}, \mathrm{proc}\}$ has its own action space $\mathcal{A}^{(m)}$ (e.g., \texttt{ADD}, \texttt{UPDATE}, \texttt{MERGE}) and is processed by the same LLM with action-specific prompts.

\subsection{Training a Memory Agent}
\mypar{Memory agent training with an SFT warm start.}
Memory-construction and memory-augmented reasoning policies are commonly trained by first performing SFT on expert- or teacher-generated trajectories and then applying RL~\citep{membuilder2026,wang2026infmem,yu2026lazymem}.
SFT teaches valid memory operations and initializes both the RL policy and, optionally, its KL reference model, thereby restricting exploration to meaningful actions.
RL then optimizes memory decisions for downstream utility beyond imitation, including what to retain, update, discard, or retrieve.
This initialization is particularly important for smaller models: MemBuilder reports that RL without SFT underperforms SFT alone, whereas SFT followed by RL performs best.

\mypar{Memory agent training with RL.}
We follow the general problem setup of MEM1~\citep{zhou2026mem} and consider an interactive, memory-augmented agent.
In our setting, the learned policy is the memory construction policy $\pi_\theta(a_t^{\mathrm{mem}} \mid h_t, \gM_{t-1})$, whose likelihood is defined over memory actions rather than retrieval or answer-generation actions.
After the memory trajectory $\tau^{\mathrm{mem}}$ is constructed, the retriever-answering produces $\hat{y}$ and the final answer quality supplies the outcome reward.
The objective can therefore be written as KL-constrained reward maximization over memory construction rollouts:
\begin{equation}
\begin{aligned}
\max_\theta \mathcal{J}(\theta)
= \max_\theta \underset{\substack{(q,\gH) \sim \mathcal{D}, \\ \tau^{\mathrm{mem}} \sim \pi_\theta(\cdot \vert \gH)}}{\mathbb{E}} \Bigg[
    R(q,\tau^{\mathrm{mem}}) - \beta \KL\left( \pi_\theta \Vert \pi_{\mathrm{ref}} \right)
\Bigg],
\end{aligned}
\end{equation}
where $\mathcal{D}$ is the training distribution over histories and questions,
$R(q,\tau^{\mathrm{mem}})$ is a measure of rollout quality and often includes outcome and process rewards,
$\pi_{\mathrm{ref}}$ is the reference model,
and $\beta$ is a constant.

\begin{figure*}[t]
    \centering
    \includegraphics[width=1\linewidth]{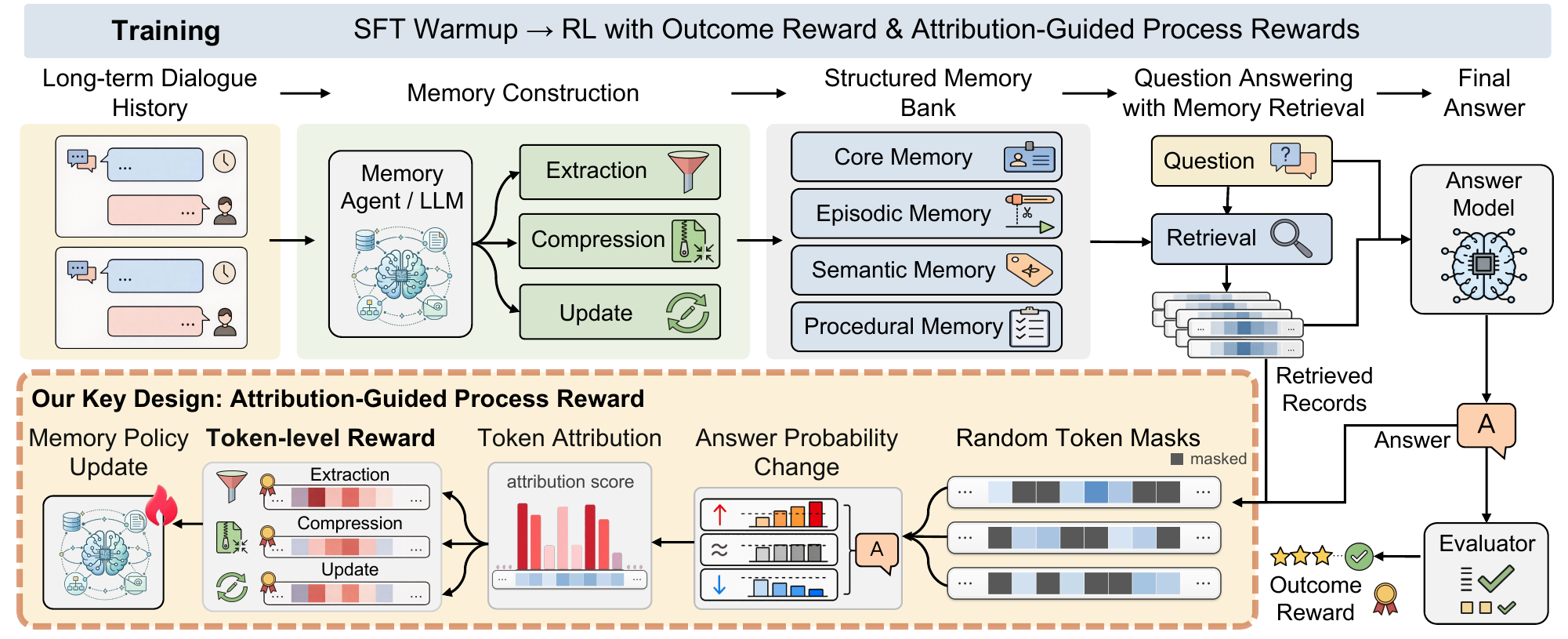}
    \caption{\textbf{Overview of AttriMem.} AttriMem constructs a structured memory bank from long-term dialogue through extraction, compression, and update, then answers questions via memory retrieval. During RL training, token-level attribution produces local process rewards, which complement the global outcome reward for fine-grained memory learning.}
    \label{method}
\end{figure*}

\mypar{Memory agent training with GRPO.}
GRPO is commonly used for RL because it avoids a separate value model~\citep{shao2024deepseekmath}.
\textit{Outcome-only} GRPO is often used for \textit{RL with verifiable reward (RLVR)}.
For each training example, the policy samples $G$ memory-construction trajectories.
A frozen retriever then retrieves from each resulting memory store, and a separate answer model (e.g., Claude) generates the final answer; neither component is part of the trained policy.
Final answer rewards are normalized within the group:
\begin{equation}
\begin{gathered}
\begin{aligned}
\max_\theta \mathcal{J}(\theta)
&= \max_{\theta} \underset{\substack{(q,\gH) \sim \mathcal{D},\\\{\tau_i^{\mathrm{mem}}\}_{i=1}^{G} \sim\pi_\mathrm{old}(\cdot \vert \gH)}}{\mathbb{E}}
\Bigg[
\frac{1}{G} \sum_{i=1}^{G} \frac{1}{T}\sum_{t=1}^{T} 
 \min\left(
    \rho_{i,t}\hat{A}^{\mathrm{out}}_{i},
    \mathrm{clip}\left( \rho_{i,t}, 1-\epsilon, 1+\epsilon \right)\hat{A}^{\mathrm{out}}_{i}
\right)- \beta \KL\left( \pi_\theta \Vert \pi_{\mathrm{ref}} \right)
\Bigg],\\
\end{aligned} \\
\text{with}~
\rho_{i,t} \coloneqq
\frac{\pi_\theta(a_{i,t}^{\mathrm{mem}} \vert h_{i,t}, \gM_{i,t-1})}
{\pi_\mathrm{old}(a_{i,t}^{\mathrm{mem}} \vert h_{i,t}, \gM_{i,t-1})},
\hat{A}^{\mathrm{out}}_{i} =
\frac{R_i^{\mathrm{out}} - \mathrm{mean}\left(\{R_j^{\mathrm{out}}\}_{j=1}^{G}\right)}
{\mathrm{std}\left(\{R_j^{\mathrm{out}}\}_{j=1}^{G}\right)},
\end{gathered}
\end{equation}
where $G$ is group size and $\pi_\mathrm{old}$ is the old policy model.
Under this formulation, GRPO with only outcome reward applies a single feedback signal to account for the quality of the entire trajectory.
This reward sparsity often results in training inefficiency and can fail to penalize incorrect intermediate actions as long as the outcome is correct.

\mypar{Context attribution.}
We formulate context attribution following \citet{cohen2024contextcite}.
Given a context $c$ and response $y \in \mathcal{Y}$, we partition the context into a set of sources for attribution: $\mathcal{Z} = \{ z_i \}_{i=1}^{\vert \mathcal{Z} \vert}$.
An attribution method is defined as $\phi: \mathcal{Z} \times \mathcal{Y} \to \R^{\vert \mathcal{Z} \vert}$, where $\phi(z_i, y ; c)$ quantifies the importance (or contribution) of source $z_i$ on response $y$ under full context $c$.
Conceptually, the attribution score corresponds to the change in response score when a source is removed:
\begin{equation}
    \phi(z_i, y;c) \approx F(y \mid c) - F(y \mid c \setminus z_i),
\end{equation}
where $F(y \mid c)$ denotes a measurement (e.g., model output logits) of response $y$ under context $c$.


\section{Method}
\label{sec:method}

\subsection{Problem Formulation}
We study long-term dialogue question answering with an explicit memory construction stage.
Given a dialogue history $\gH = \{ h_t \}_{t=1}^{T}$ and a question $q$ asked after the history, the system first constructs a memory store $\gM_T$ from $\gH$ and then answers $q$ using a fixed retrieval-and-answering interface over $\gM_T$.
In this paper, \ourshort{} focuses on learning the \textit{memory construction policy}:
What information should be extracted, updated, merged, compressed, and retained in the memory store?
Thus, performance improvements should be attributed to better memory construction; the learned policy produces memory records that are more complete, less noisy, and more useful for later retrieval, reasoning, and question answering.
The retriever and answer generator are used for question answering, but they are not the target of our method.

Formally, at each session $t$, a memory construction policy $\policy$ observes the current session $h_t$ and the previous memory store $\gM_{t-1}$, and produces a memory-construction action $a_t^{\mathrm{mem}}$.
The action updates the memory store as
\begin{equation}
    \gM_t = U(\gM_{t-1}, h_t, a_t^{\mathrm{mem}}),
\end{equation}
where $U(\cdot)$ denotes the memory update operator.
Let $\Delta \gM_t$ be the change in memory content, yielding a memory-construction trajectory $\tau^{\mathrm{mem}} = \{(h_t, \gM_{t-1}, a_t^{\mathrm{mem}}, \Delta \gM_t)\}_{t=1}^{T}$.
After processing all sessions, a retriever-answering interface produces answer
\begin{equation}
    \hat{y} = G\big(q, \mathrm{Ret}(q, \gM_T)\big),
\end{equation}
where $G(\cdot)$ denotes the answer generator.
The training signal is derived from the final answer quality, while the optimization targets are the memory-construction actions $\{ a_t^{\mathrm{mem}} \}_{t=1}^{T}$.
This formulation isolates the central question of the paper:
\textit{How can we train a model to build a better memory store?}

\begin{center}
\begin{mainbox}
\textbf{Why attribution is needed: }Outcome can tell whether the final answer is correct, but it cannot tell which memory action caused the success or failure.
This is especially problematic for long dialogue histories, where many memory records are irrelevant, redundant, or only partially useful.
\textbf{Attribution-guided reward adds information that is absent from the final score}: it identifies which constructed memory content actually supports the answer.
Based on this, \ourshort{} changes the credit-assignment signal for memory construction.
\end{mainbox}
\end{center}

\subsection{Attribution-based Process Reward}
\label{sec:method:attribution}

To address this fine-grained credit-assignment bottleneck, we assign process rewards to individual tokens within intermediate memory actions.

\mypar{Decomposed policy gradient.}
Our design of process rewards optimizes the memory construction policy.
Let $a_{i,t}^{\mathrm{mem}}=(a_{i,t,1}^{\mathrm{mem}},\ldots,a_{i,t,L_{i,t}}^{\mathrm{mem}})$ denote the token sequence of the $t$-th memory-construction action in rollout $i$.
Based on the GRPO formulation, the policy gradient over action tokens is as follows (ignoring the KL penalty term for simplicity):
\begin{equation}
\begin{gathered}
\begin{aligned}
\nabla_\theta \mathcal{J}(\theta) & = \underset{\substack{(q,\gH) \sim \mathcal{D},\\ \{\tau_i^{\mathrm{mem}}\}_{i=1}^{G} \sim\pi_\mathrm{old}(\cdot \vert \gH)}}{\mathbb{E}}
\Bigg[ \frac{1}{G} \sum_{i=1}^{G}\frac{1}{T} \sum_{t=1}^{T}\frac{1}{L_{i,t}}\sum_{k=1}^{L_{i,t}} 
    \rho_{i,t,k}\hat{A}_{i,t,k}
    \nabla_\theta \log \pi_\theta(a_{i,t,k}^{\mathrm{mem}} \vert h_{i,t}, \gM_{i,t-1}, a_{i,t,<k}^{\mathrm{mem}})
    \Bigg],
\end{aligned}\\
~\text{where}~
\rho_{i,t,k} \coloneqq
\frac{\pi_\theta(a_{i,t,k}^{\mathrm{mem}} \vert h_{i,t}, \gM_{i,t-1}, a_{i,t,<k}^{\mathrm{mem}})}
{\pi_\mathrm{old}(a_{i,t,k}^{\mathrm{mem}} \vert h_{i,t}, \gM_{i,t-1}, a_{i,t,<k}^{\mathrm{mem}})},
\hat{A}_{i,t,k} \coloneqq
\hat{A}^{\mathrm{out}}_i + \lambda \hat{A}^{\mathrm{proc}}_{i,t,k}.
\end{gathered}
\end{equation}
where $\hat{A}^{\mathrm{out}}_i$ is the within-group outcome advantage,
$\hat{A}^{\mathrm{proc}}_{i,t,k}$ is the token-level process advantage derived from $r_{i,t,k}^{\mathrm{proc}}$ according to GRPO advantage estimation,
and $\lambda$ controls the weight of the process signal.
The outcome advantage broadcasts final-answer quality to all action tokens in the trajectory, as in outcome-only GRPO, while the process advantage gives distinct attribution-based credit to each action token.

\mypar{Measuring the contribution of action tokens to the final answer.}
The \textbf{intuition} is to estimate the counterfactual contribution of intermediate steps:
\textit{if a source (e.g., a token or a sequence of tokens) were not present in the trajectory, how much would the final outcome change?}

Let $\Delta \gM_t$ denote the textual memory content produced or modified by $a_t^{\mathrm{mem}}$ when updating $\gM_{t-1}$ to $\gM_t$.
To estimate the contribution of the $k$-th token in the $t$-th memory action, we treat $a_{t,k}^{\mathrm{mem}}$ as an attribution source and measure how counterfactually removing that token from the constructed memory context changes the score of the generated answer $\hat{y}$.
Specifically, we compute $\phi(a_{t,k}^{\mathrm{mem}},\hat{y};c)$, where $\phi$ is an attribution method and $c$ is the full attribution context.

In this paper, we adopt \textit{ContextCite}~\citep{cohen2024contextcite}, which fits a lightweight linear attribution model by randomly ablating contexts.
Specifically, it approximates the response score as a linear function of whether a context source is retained.
We evaluate up to 32 randomly ablated variants in parallel, reducing the additional wall-clock latency to approximately one batched forward pass.
We defer implementation details to Appendix~\ref{sec:appendix:impl}.

\mypar{Computing attribution rewards.}
We use each token's attribution score directly as its process reward:
\begin{equation}
r_{i,t,k}^{\mathrm{proc}}
=\phi(a_{i,t,k}^{\mathrm{mem}},\hat{y}_i;c_i).
\end{equation}
Thus, tokens within the same action receive distinct signed rewards rather than sharing an aggregated action-level reward.


\section{Experiments}
\subsection{Experimental Setup}
\label{sec:setup}

\mypar{Datasets and metrics.}
We train on LongMemEval~\citep{wu2024longmemeval} and evaluate transfer on two additional benchmarks, LoCoMo~\citep{maharana2024locomo} and PerLTQA~\citep{du2024perltqa}.
The two benchmarks are used for out-of-domain evaluation under a zero-shot transfer setting, enabling assessment of both in-domain memory policy learning and cross-domain generalization. We report accuracy as the primary evaluation metric across all benchmarks.

\vspace{1mm}
\mypar{Baselines.}
We compare \ourshort{} against three categories of baselines.
Training-free retrieval methods include RAG-Session and RAG-Utterance~\citep{lewis2020rag}, which directly retrieve historical context at the session or utterance level.
Heuristic-based memory methods include Mem0~\citep{mem02024}, MIRIX~\citep{mirix}, MemoryOS~\citep{memoryos}, A-Mem~\citep{amem}, LightMem~\citep{lightmem2026},and GAM~\citep{gam}.
RL-based memory methods include Memory-R1~\citep{yan2026memory}, Mem-T~\citep{memt2026}, and MemBuilder~\citep{membuilder2026}.
For MemBuilder, our closest prior work, and \ourshort{}, we report base, +SFT, +RL, and +SFT+RL variants to isolate the effects of SFT and RL, both separately and in combination, and compare their reward designs from comparable supervised initializations.

\vspace{1mm}
\mypar{Implementation details.}
Since MemBuilder~\citep{membuilder2026} is the most closely related recent work, we follow its backbone and memory architecture to enable a controlled comparison.
Specifically, both MemBuilder and \ourshort{} use Qwen3-4B and undergo SFT warmup followed by GRPO-based RL~\citep{shao2024deepseekmath}.
We keep the SFT initialization, optimization budget, shared hyperparameters, retrieval-and-answering interface, decoding configuration, and evaluation protocol consistent.
We use 3,000 SFT steps and 400 RL steps, with detailed hyperparameters provided in Appendix~\ref{sec:appendix:impl}.

\begin{table}[t]
\centering
\small
\resizebox{0.56\columnwidth}{!}{
\renewcommand{\arraystretch}{0.9}
\begin{threeparttable}
\begin{tabular}{lccc}
\toprule
\textbf{Method} & \textbf{LoCoMo} & \textbf{LongMemEval} & \textbf{PerLTQA} \\
\midrule

\multicolumn{4}{l}{\textit{Training-Free Retrieval}} \\
RAG-Session$^\dagger$   & 70.35 & 66.75 & 79.21 \\
RAG-Utterance$^\dagger$ & 74.87 & 69.00 & 77.23 \\
\midrule

\multicolumn{4}{l}{\textit{Heuristic Memory Construction}} \\
Mem0$^\dagger$                         & 51.64 & 47.00 & 62.04 \\
MIRIX$^\dagger$                        & 77.48 & 73.25 & 83.11 \\
MemoryOS   & 52.92 & 57.75 & 64.08 \\
A-Mem & 68.98 & 60.00 & 74.66 \\
LightMem & 80.53 & 63.50 & 80.30\\
GAM & 79.50 & 73.25 & 81.59 \\
\midrule

\multicolumn{4}{l}{\textit{RL-based Memory Learning}} \\
Memory-R1$^\dagger$        & 62.74 & -- & -- \\
Mem-T            & 75.67 & 69.75 & 76.58 \\
MemBuilder (Qwen3-4B)      & 69.74 & 56.00 & 78.33 \\
\quad + SFT                & 78.70 & 80.50 & 82.36 \\
\quad + RL                 & 69.84 & 58.00 & 79.41 \\
\quad + SFT + RL           & 80.01 & 81.75 & 83.74 \\
\textbf{Ours} (Qwen3-4B)     & 67.88 & 55.75 & 78.32 \\
\quad + SFT                & 78.40 & 80.25 & 81.52 \\
\quad + \textbf{RL$_{Tok}$}                 & 72.05 & 61.25 & 80.58 \\
\quad + SFT + \textbf{RL$_{Tok}$}           & \textbf{82.48} & \textbf{83.25} & \textbf{84.49} \\
\bottomrule
\end{tabular}

\end{threeparttable}
}
\caption{
\textbf{Overall performance comparison.}
Results marked with $^\dagger$ are reported by \citet{membuilder2026}.
}
\label{tab:memory_construction_results}
\end{table}

\begin{figure*}[t]
\centering
\includegraphics[
  width=\textwidth,
  trim=10 25 10 25,
  clip
]{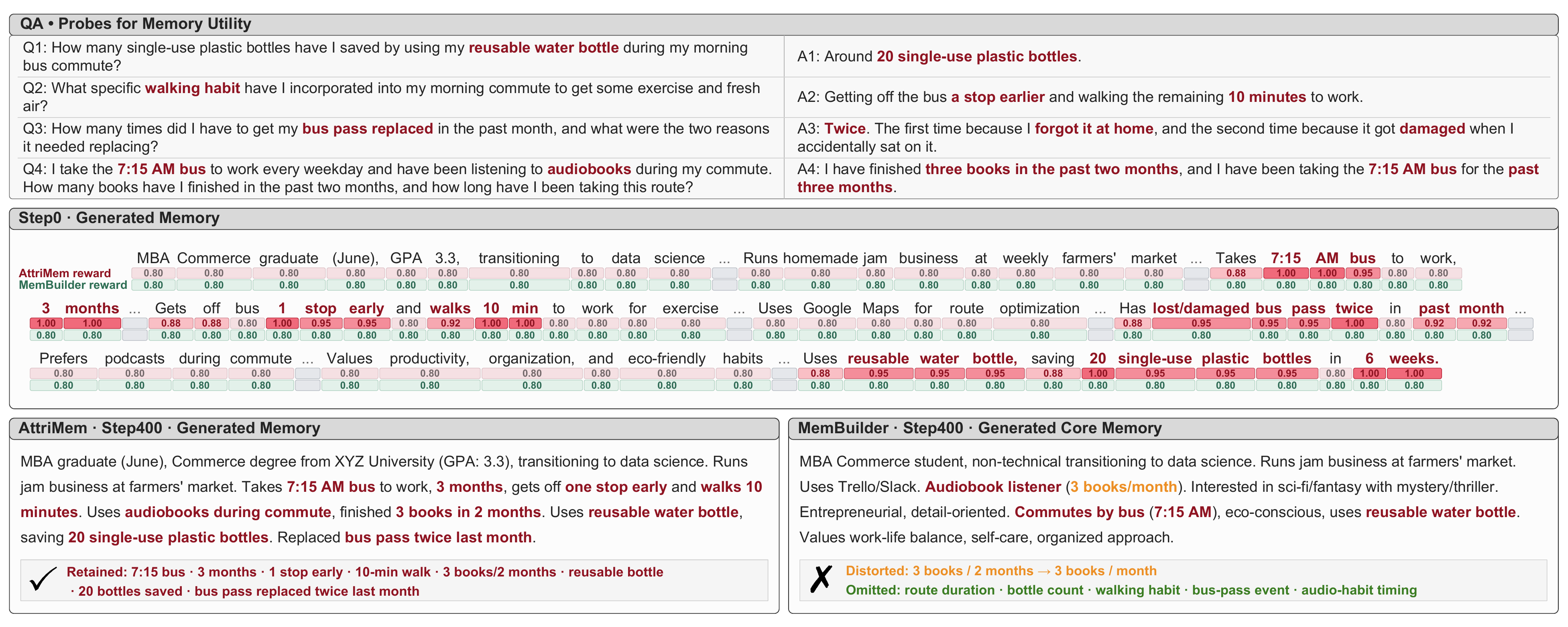}
\caption{\textbf{Case study of attribution-guided memory.}
Before RL, MemBuilder evaluates the whole output, whereas AttriMem distinguishes answer-critical evidence from distracting content through token-level credit. After RL, AttriMem retains critical evidence and removes distracting information, whereas the MemBuilder-trained policy omits and distorts answer-critical details.}
\label{fig:reward_case}
\end{figure*}

\subsection{Main Results}
\label{sec:results}

\mypar{Overall performance.}
\label{sec:results:locomo}
We compare \ourshort{} with different baselines. As shown in Table~\ref{tab:memory_construction_results}, \textit{\textbf{the fine-grained attribution-based reward in \ourshort{} is consistently effective, with its advantage over MemBuilder emerging during RL.}} Starting from comparable SFT performance, \ourshort{} reaches 82.48, 83.25, and 84.49 after RL, outperforming MemBuilder by 2.47, 1.50, and 0.75 points on the three benchmarks, respectively. \textit{\textbf{\ourshort{} achieves the best performance across all three benchmarks.}} It surpasses the strongest prior methods by 1.95 points on LoCoMo, 1.50 points on LongMemEval, and 0.75 points on PerLTQA. \textit{\textbf{\ourshort{} also transfers effectively to out-of-domain benchmarks.}} Although trained only on LongMemEval, it reaches 82.48 on LoCoMo and 84.49 on PerLTQA, exceeding the strongest non-RL baselines.

\vspace{1mm}
\mypar{Memory utility across answer models.}
Because the memory policy is optimized using feedback from a specific answer model, we examine whether the resulting memory remains useful with other answer models. \textit{\textbf{The learned memory can be directly reused by different answer models.}} As shown in Table~\ref{tab:answer_model_comparison}, Claude, GPT, and Qwen3 can all effectively use the memory, indicating that their utility is not tied to a particular answer model. \textit{\textbf{Memory learning also improves the backbone's ability to reason over retrieved evidence.}} When the \ourshort{}-trained Qwen3-4B is used as the answer model, it improves over the base model by 0.95, 13.75, and 2.15.

\begin{table}[t]
\centering
\small
\renewcommand{\arraystretch}{0.7}
\resizebox{0.7\columnwidth}{!}{
\begin{tabular}{lccc}
\toprule
\textbf{Answer Model} & \textbf{LoCoMo} & \textbf{LongMemEval} & \textbf{PerLTQA} \\
\midrule
Claude 4.5 Sonnet & 82.48 & 83.25 & 84.49 \\
GPT-4.1           & 80.41 & 82.00 & 82.18 \\
Qwen3-4B (Base)     & 76.69 & 65.00 & 80.46 \\
Qwen3-4B (AttriMem-trained)     & 77.64 & 78.75 & 82.61 \\
\bottomrule
\end{tabular}
}
\caption{Performance on different answer models.}
\label{tab:answer_model_comparison}
\end{table}

\vspace{1mm}
\mypar{Case study.}
To better understand why \ourshort{} is effective, we compare MemBuilder and \ourshort{} on the same pre-RL output and examine how their outputs differ after RL.
\textbf{\emph{\ourshort{} benefits from token-level credit assignment, which selectively reinforces answer-critical details within each memory output.}}
As shown in Figure~\ref{fig:reward_case}, MemBuilder assigns one reward to the entire output, whereas \ourshort{} assigns distinct credit to exact times, durations, and quantities.
This finer-grained signal tells the policy which specific tokens are more important, whereas MemBuilder broadcasts the same reward to every token and thus cannot distinguish useful facts.



\subsection{Ablation Study}
\label{sec:analysis:ablation}

\mypar{Effect of reward granularity.}
We compare outcome-only RL (\textit{RL$_{Out}$}), action-level attribution (\textit{RL$_{Act}$}), and token-level attribution (\textit{RL$_{Tok}$}). As shown in Table~\ref{tab:reward_granularity_ablation}, \textbf{\textit{attribution-guided rewards consistently improve performance, with token-level credit yielding the strongest results.}} Without SFT, \textit{RL$_{Tok}$} improves the base model by 4.17, 5.50, and 2.26 points on LoCoMo, LongMemEval, and PerLTQA, outperforming \textit{RL$_{Act}$} by 0.95, 1.50, and 0.60 points. The same trend holds with SFT, where \textit{SFT+RL$_{Tok}$} achieves the best results.

\begin{table}[t]
\centering
\small
\setlength{\tabcolsep}{8pt}
\renewcommand{\arraystretch}{0.7}
\begin{threeparttable}
\resizebox{0.6\columnwidth}{!}{
\begin{tabular}{lccc}
\toprule
\textbf{Method} & \textbf{LoCoMo} & \textbf{LongMemEval} & \textbf{PerLTQA} \\
\midrule
Ours (Qwen3-4B)
& 67.88  & 55.75 & 78.32 \\

\quad + RL$_{Out}$
& 70.19 & 58.25 & 79.61 \\

\quad + RL$_{Act}$
& 71.10 & 59.75 & 79.98 \\

\quad + \textbf{RL$_{Tok}$}
& 72.05 & 61.25 & 80.58 \\

\quad + SFT + RL$_{Out}$
& 79.46 & 81.50 & 83.69 \\

\quad + SFT + RL$_{Act}$
& 81.32 & 82.00 & 84.05 \\

\quad + SFT + \textbf{RL$_{Tok}$}
& \textbf{82.48} & \textbf{83.25} & \textbf{84.49} \\
\bottomrule
\end{tabular}
}
\end{threeparttable}
\caption{Ablation study on reward granularity.}
\label{tab:reward_granularity_ablation}
\end{table}



\vspace{1mm}
\mypar{RL beyond SFT initialization.}
We examine the effectiveness of attribution-guided RL with and without SFT initialization. As shown in Table~\ref{tab:memory_construction_results}, \textbf{\textit{\ourshort{} consistently improves performance with or without SFT, whereas MemBuilder yields only limited gains when trained directly from the base model.}} Without SFT, \textit{RL$_{Tok}$} improves the base model by 2.26--5.50 points, compared with 0.10--2.00 points for MemBuilder. This suggests that coarse-reward RL struggles to achieve meaningful gains without SFT, whereas \textbf{\textit{\ourshort{} provides fine-grained reward that reduces reliance on SFT.}}

\subsection{Further Analysis}
\label{sec:analysis:quality}

\mypar{Intermediate memory quality.}
We use a GPT-based judge to compare intermediate memory, with details provided in Appendix~\ref{LLM_judge_protocol}. As shown in Table~\ref{tab:intermediate_rationality}, \textbf{\textit{token-level attribution improves the intermediate memory process itself.}} Across LoCoMo, LongMemEval, and PerLTQA, Ours$_{Tok}$ achieves win rates of 73.91\%, 72.94\%, and 77.78\% over MemBuilder, respectively. It is also consistently preferred over Ours$_{Out}$ and Ours$_{Act}$, with win rates of 61.11--71.43\% and 58.97--61.80\%.

\begin{figure}[!t]
\centering
\includegraphics[width=.5\linewidth]{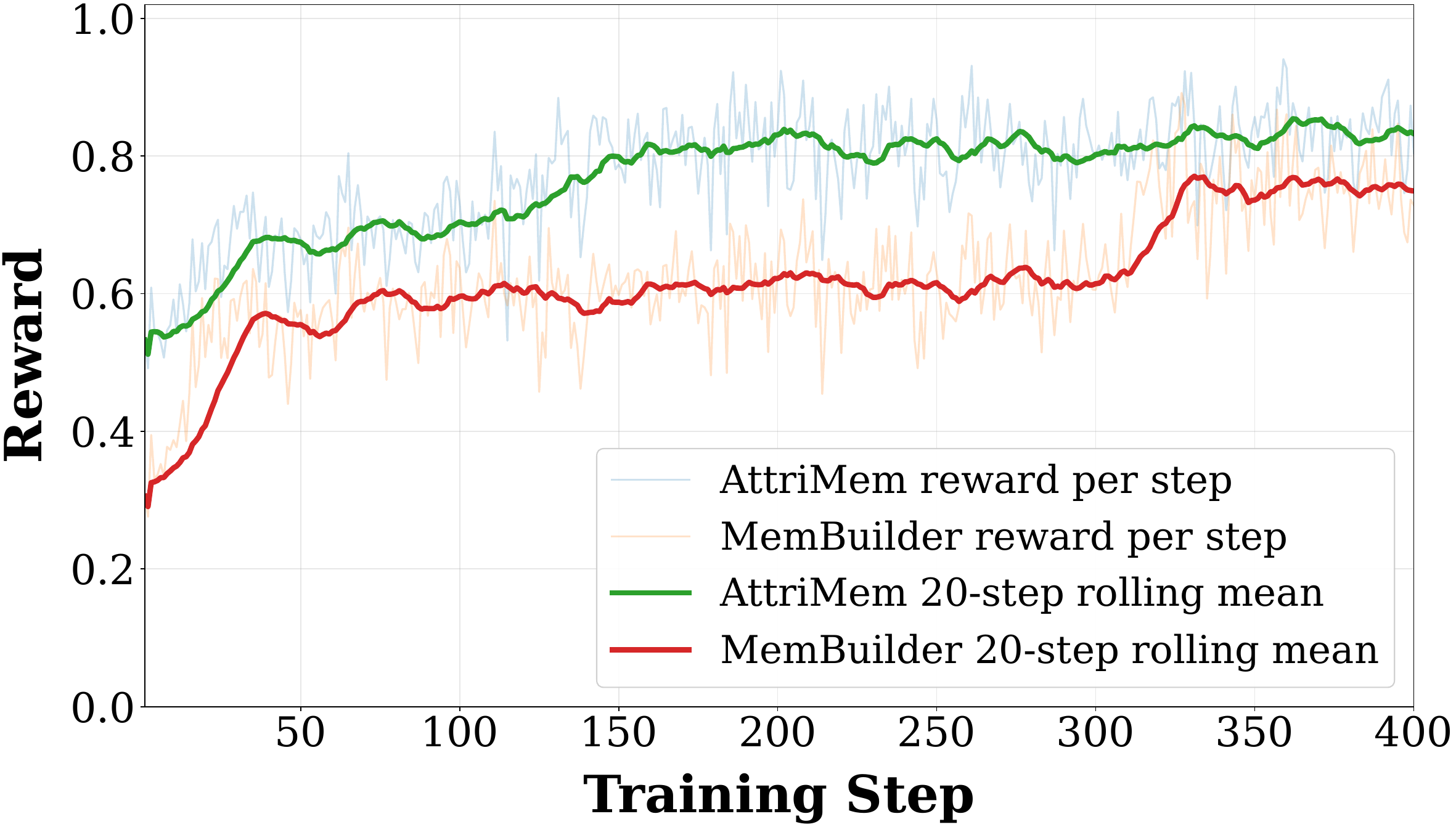}
\caption{Reward evolution during RL training. }
\label{fig:training_dynamics}
\end{figure}

\vspace{1mm}
\mypar{RL training dynamics.}
We track reward evolution during RL training. As shown in Figure~\ref{fig:training_dynamics}, \textbf{\textit{token-level attribution provides a more stable and steadily improving optimization signal than aggregated rewards.}}
\ourshort{} rises consistently, reaches a high-reward region midway through training, and remains stable. In contrast, MemBuilder plateaus after early gains and improves only through a sharp surge around step 320, indicating delayed and less stable optimization.

\begin{table}[t]
\centering
\small
\setlength{\tabcolsep}{8pt}
\renewcommand{\arraystretch}{0.7}
\resizebox{0.7\columnwidth}{!}{
\begin{threeparttable}

\begin{tabular}{lccc}
\toprule
\textbf{Comparison} & \textbf{LoCoMo} & \textbf{LongMemEval} & \textbf{PerLTQA} \\
\midrule
\textbf{Ours$_{Tok}$} vs. MemBuilder
& 73.91 & 72.94 & 77.78 \\

\textbf{Ours$_{Tok}$} vs. Ours$_{Out}$
& 61.11 & 68.05 & 71.43 \\

\textbf{Ours$_{Tok}$} vs. Ours$_{Act}$
& 58.97 & 61.76 & 61.80 \\
\bottomrule
\end{tabular}
\end{threeparttable}
}
\caption{
\textbf{Evaluation of memory quality.} We report the \textbf{\textit{win rate (\%)}} of \ourshort{} against each compared method, where higher values indicate that \ourshort{} is more frequently judged to produce more reasonable intermediate memory.
}
\label{tab:intermediate_rationality}
\end{table}

\begin{figure}[t]
    \centering
    \includegraphics[width=0.5\linewidth]{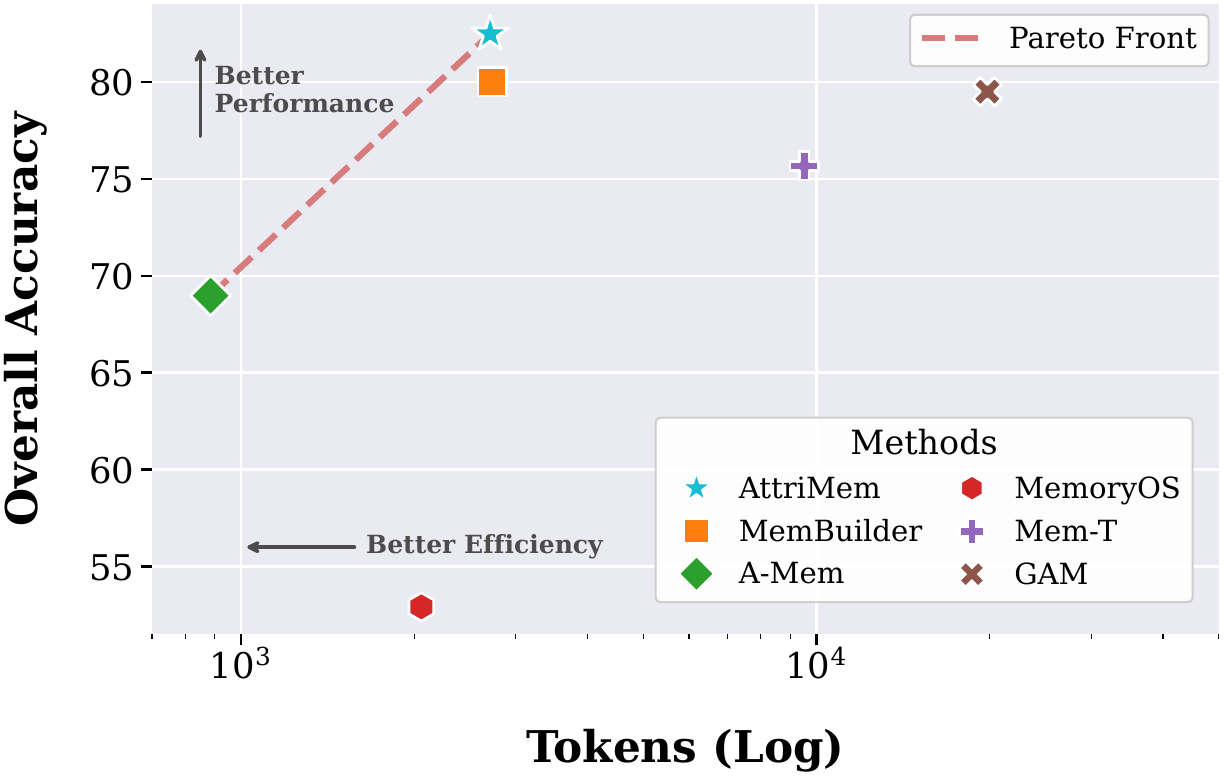}
    \caption{
    Performance--efficiency trade-off on LoCoMo.
    The horizontal axis denotes average end-to-end token consumption
    per question, and the vertical axis denotes answer accuracy.
    }
    \label{fig:token_efficiency}
\end{figure}

\mypar{Inference token efficiency.}
We report average end-to-end tokens per question, including amortized memory construction and question answering. As shown in Figure~\ref{fig:token_efficiency}, \textbf{\textit{AttriMem achieves the best performance--efficiency trade-off and lies on the Pareto frontier.}} It attains the highest accuracy with token consumption comparable to MemBuilder and substantially lower than Mem-T and GAM. Although A-Mem uses fewer tokens, its accuracy is considerably lower.


\section{Related Work}
\label{sec:related}

\mypar{RAG and heuristic memory.}
Retrieval-augmented methods retrieve relevant knowledge from external sources~\citep{lewis2020rag}, and heuristic memory methods typically use hand-crafted rules or scoring functions for writing, retaining, or compressing~\citep{gam2024,amem,lightmem2026,mem02024}. However, their memory policies are predefined, which can lead to an objective mismatch.

\vspace{1mm}
\mypar{Outcome-redistributed rewards.}
RL-based methods learn memory from task rewards, making memory task-grounded. Methods such as \memalpha~\citep{memalpha2024} optimize memory policies using task-performance-derived feedback, while \memt~\citep{memt2026} redistributes final rewards to intermediate nodes in sampled memory-action trees. However, these signals remain outcome-derived: they indicate downstream success but provide limited evidence about which memory action, content span, or token supports the final answer.

\vspace{1mm}
\mypar{QA-derived action rewards.}
The closest work to ours is \membuilder~\citep{membuilder2026}, which uses synthetic session-level QA to evaluate the memory bank and forms action-level rewards from QA performance and retrieval usage. Although this provides denser supervision than final task success alone, the reward is still too coarse to identify which action, memory entry, or content span should receive credit. In contrast, \ourshort{} derives token-level process feedback and aligns it with the specific content produced within memory action.

\section{Discussion and Future Work}
\label{sec:limitations}
\mypar{Computational overhead.}
Estimating token contributions requires evaluating randomly masked variants, introducing additional computation during training. However, the additional latency remains limited because all masked variants can be evaluated within a single batched model forward pass.

\vspace{1mm}
\mypar{Attribution quality.}
\ourshort{} relies on the quality of token-level attribution. For example, in exceptionally long contexts, attribution can be more challenging. However, in our experiments, LongMemEval is itself an exceptionally long-context benchmark, and our experiments on it suggest that this issue is limited. Another potential source of attribution noise is the use of weaker answer models. However, in practice, modern agent systems typically employ capable answer models.

Moreover, attribution-based process rewards also complement rather than replace the global outcome reward, consistently improving the corresponding RL variants without observed regressions. More importantly, these concerns primarily stem from the attribution estimator rather than \ourshort{} itself: \ourshort{} is not tied to ContextCite and can adopt more robust long-context or white-box methods.

\vspace{1mm}
\mypar{Token-representable memory.}
Our current framework assumes that intermediate memory-action outputs can be decomposed into tokens and aligned with attribution scores. However, whether tokens are the optimal attribution unit remains open, as answer-critical evidence may span multiple tokens or depend on interactions among them.


\section{Conclusion}

We presented \ourshort{}, an attribution-guided process-feedback framework for RL-based memory learning. \ourshort{} addresses the fine-grained credit-assignment bottleneck in existing memory-learning methods by tracing final-answer support back to intermediate memory contents and converting token-level contributions into local process rewards. Experiments on long-horizon dialogue question answering show that attribution-guided process feedback provides more precise and effective supervision than outcome-only or action-level rewards, improving overall performance, intermediate memory quality, and RL training stability.



\bibliographystyle{assets/acl_natbib}
\bibliography{references}


\clearpage
\beginappendix

\begin{table}[!t]
\begin{center}
\footnotesize
\begin{tabular}{@{}lp{0.75\linewidth}@{}}
\toprule
Symbol & Meaning \\
\midrule
$a_t^{\mathrm{mem}}$ & Memory-construction action token sequence at session $t$. \\
$a_{t,k}^{\mathrm{mem}}$ & Token $k$ in memory-construction action $a_t^{\mathrm{mem}}$. \\
$L_t$ & Number of tokens in memory-construction action $a_t^{\mathrm{mem}}$. \\
$h_t \in \gH$ & Session $t$ in the conversation history. \\
$\tau^{\mathrm{mem}}$ & Memory-construction trajectory. \\
$\gM_t$ & Memory store after processing session $t$. \\
$\Delta \gM_t$ & \makecell[l]{Textual memory content produced\\or modified by $a_t^{\mathrm{mem}}$.} \\
$\pi_\theta$ & Agent policy parameterized by $\theta$. \\
$G$ & GRPO group size. \\
$q \in \mathcal{Q}$ & Query from query set. \\
$r_{t,k}^{\mathrm{proc}}$ & Process reward for action token $a_{t,k}^{\mathrm{mem}}$. \\
$\phi(z_i,y;c)$ & \makecell[l]{Attributed contribution of source $z_i$ on\\response $y$ under full context $c$.} \\
\bottomrule
\end{tabular}    
\end{center}
\caption{Summary of notations.}
\end{table}

\section{ContextCite Methodology Details}
\label{sec:appendix:contextcite}

We instantiate the attribution function $\phi$ with ContextCite~\citep{cohen2024contextcite}, adapting its source-level formulation to the action-token outputs used by \ourshort{}.
ContextCite permits a source to be any subset of context tokens and also studies fine-grained word-level sources; here, we use individual model tokens as sources.
For a generated answer $\hat{y}$, let the attribution context be $c = (z_1,\ldots,z_d)$, where each source $z_j$ corresponds to a generated action-token position $a_{t,k}^{\mathrm{mem}}$ in the constructed textual memory context.
The objective is to estimate how retaining each source changes the answer model's score for the fixed answer $\hat{y}$.

\mypar{Context ablations.}
We represent an ablation by a binary inclusion vector $v \in \{0,1\}^{d}$, where $v_j=1$ indicates that the corresponding action token is retained and $v_j=0$ indicates that it is removed.
Let $\operatorname{Ablate}(c,v)$ denote the counterfactual memory context formed by retaining the selected action tokens in their original order.
Following ContextCite, we independently retain each source with probability $1/2$, which is equivalent to sampling $v$ uniformly from $\{0,1\}^{d}$.
For $N$ sampled masks $\{v_k\}_{k=1}^{N}$, the corresponding ablated contexts can be evaluated in one batch.

\mypar{Scoring the fixed answer.}
Let $p_{\mathrm{ans}}$ denote the fixed answer model used to score $\hat{y}=(\hat{y}_1,\ldots,\hat{y}_L)$.
For each ablation vector, we evaluate the probability of the same generated answer under teacher forcing:
\begin{equation}
\begin{aligned}
f(v_k)
&\coloneqq p_{\mathrm{ans}}\!\left(\hat{y}\mid q,
\operatorname{Ablate}(c,v_k)\right) \\
&= \prod_{\ell=1}^{L}
p_{\mathrm{ans}}\!\left(\hat{y}_{\ell}\mid q,
\operatorname{Ablate}(c,v_k),\hat{y}_{<\ell}\right).
\end{aligned}
\end{equation}
As in ContextCite, we regress on the logit-transformed probability
\begin{equation}
g(v_k) \coloneqq \sigma^{-1}(f(v_k))
= \log \frac{f(v_k)}{1-f(v_k)},
\end{equation}
rather than on the bounded probability itself.
Holding $\hat{y}$ fixed across ablations isolates how the included memory sources affect the model's preference for the observed answer; it does not resample an answer for each ablated context.

\mypar{Sparse linear surrogate.}
ContextCite approximates $g(v)$ with a linear surrogate $\hat{g}(v)=b+w^{\top}v$.
Given the ablation dataset $\{(v_k,g(v_k))\}_{k=1}^{N}$, we fit the surrogate with LASSO:
\begin{equation}
(\hat{b},\hat{w})
= \argmin_{b,w}
\frac{1}{2N}\sum_{k=1}^{N}
\left(g(v_k)-b-w^{\top}v_k\right)^2
+ \alpha\lVert w\rVert_1.
\end{equation}
The sparsity penalty reflects the assumption that only a small subset of the context materially affects a particular answer and allows the surrogate to be estimated from substantially fewer masks than sources.
Because the surrogate is linear in the inclusion indicators, its fitted coefficient for source $z_j$ is used directly as the attribution score:
\begin{equation}
\phi(z_j,\hat{y};c) \coloneqq \hat{w}_j.
\end{equation}
A positive coefficient means that retaining the source increases the score of $\hat{y}$ under the local surrogate, whereas a negative coefficient indicates that retaining it decreases the score.

\mypar{Mapping attribution to process rewards.}
Each source index $j$ maps directly to one generated policy token $a_{t,k}^{\mathrm{mem}}$.
We use its ContextCite coefficient directly as the token-level process reward:
\begin{equation}
r_{t,k}^{\mathrm{proc}}
=\phi(a_{t,k}^{\mathrm{mem}},\hat{y};c)
=\hat{w}_j.
\end{equation}
This reward is applied only to the corresponding action token in \cref{sec:method:attribution}; no action-level aggregation or length normalization is used.

\mypar{Ablation budget.}
The original ContextCite experiments use 32 independently sampled ablations and LASSO regularization $\alpha=0.01$.
We use the same maximum ablation budget and increase the cumulative number of masks from 8 to 16 and then 32 at RL steps 0, 80, and 200, respectively.
This schedule reduces attribution cost early in training while improving the fidelity of process rewards as the memory policy becomes stronger.

\section{Implementation Details}
\label{sec:appendix:impl}

\mypar{Memory policy and fixed QA interface.}
We use Qwen3-4B as the memory-construction policy for all \ourshort{} variants.
At session $t$, the policy receives the current session $h_t$ and the previous memory store $\gM_{t-1}$ and produces a memory action $a_t^{\mathrm{mem}}$.
The memory store contains core, episodic, semantic, and procedural actions; the same policy backbone is used with action-specific prompts and action spaces.
After all sessions have been processed, a frozen retriever selects information from $\gM_T$ and a separate answer model generates $\hat{y}$.
The retriever and answer model are used to obtain training feedback but are not optimized as part of the memory policy.

\mypar{Two-stage training.}
We train the memory policy on LongMemEval in two stages.
The SFT stage provides the initialization used by the SFT and SFT+RL variants and runs for 10 epochs (3,000 update steps), with learning rate $5\times10^{-7}$ and batch size 32.
Starting from either the base model or the SFT initialization, the RL stage applies GRPO for 400 update steps with learning rate $1\times10^{-6}$, effective batch size 256, group size $G=8$, and maximum sequence length 6,000.
The GRPO objective includes KL regularization against the reference policy.
For each training example, the old policy samples $G$ memory-construction trajectories; the fixed retrieval-and-answering interface evaluates each resulting memory store, and the answer rewards are normalized within the group.
Each action token's advantage combines the resulting trajectory-level outcome advantage with its attribution-based token-level process advantage defined in \cref{sec:method:attribution}.

\mypar{Attribution computation.}
We provide the complete ContextCite formulation and its token-level adaptation in Appendix~\ref{sec:appendix:contextcite}.
For each generated answer, we keep the answer fixed while evaluating randomly ablated contexts, fit the sparse linear surrogate, and map each coefficient directly back to the corresponding generated action token.
The cumulative ablation budget increases from 8 to 16 and then 32 masks at RL steps 0, 80, and 200, respectively.
All masks for an attribution estimate are evaluated in parallel as a batch.


\mypar{Controlled comparison and evaluation.}
For a consistent comparison, we standardize the memory-construction model to Qwen3-4B for all baselines that allow model replacement. For training-dependent methods whose learned parameters are integral to the original approach, such as Mem-T, we instead use the officially released models trained under their original settings. Across all methods, Claude 4.5 Sonnet serves as the shared answer model, and GPT-4.1 serves as the shared judge model. For the closest controlled comparison with MemBuilder~\citep{membuilder2026}, we further match the memory architecture, SFT initialization, optimization budget, shared hyperparameters, retrieval-and-answering interface, decoding configuration, and evaluation protocol. The designed difference is the RL reward: \ourshort{} supplements the outcome reward with attribution-based token-level process rewards. We train on LongMemEval and evaluate zero-shot transfer to LoCoMo and PerLTQA, reporting accuracy on all three benchmarks.
The answer-model comparison additionally evaluates the learned memory with Claude 4.5 Sonnet, GPT-4.1, and Qwen3-4B, as reported in \cref{tab:answer_model_comparison}.

\mypar{LLM judge protocol (for \cref{sec:analysis:quality}).}
\label{LLM_judge_protocol}
The LLM judge receives the following context for each pairwise comparison:
(1) the full original dialogue, (2) the question, (3) the gold answer,
(4) the intermediate outputs of System A and System B at the specified
stage (extraction / compression / retrieval).
Method identities are replaced with ``System A'' and ``System B''.
Each pair is evaluated twice (order A-then-B and B-then-A); the final
win rate averages both orderings.
The actual prompt is shown in \cref{fig:memory_construction_evaluation_prompt}.

\begin{figure*}[t]
\centering
\begin{tcolorbox}[
    width=\textwidth,
    left=1.5mm,
    right=1.5mm,
    top=1.5mm,
    bottom=1.5mm,
    colback=black!5!white,
    colframe=black,
    title={Prompt template for evaluating intermediate memory-construction quality.}
]
\begin{lstlisting}[
    basicstyle=\tiny\ttfamily,
    columns=flexible,
    breaklines=true
]
You are evaluating intermediate memory-construction quality for a long-term memory QA system.

Given:
1. A conversation session.
2. Downstream QA targets that the memory system should eventually help answer.
3. Two systems' intermediate memory construction outputs for the same session. The outputs may contain one memory module or all memory modules together.

Your job is to decide which system produced the more evidence-grounded, useful, and correct memory construction output for this step.

Use this priority order:
1. Evidence-grounded faithfulness: prefer memory updates that are directly traceable to concrete evidence in the conversation session. Penalize hallucinated facts, wrong dates, wrong speakers, unsupported causal claims, and plausible but ungrounded details.
2. Answer-critical coverage: prefer memories that preserve facts needed to answer the downstream questions, including entities, dates, counts, relationships, preferences, events, objects, motivations, and reasons.
3. Operation rationality: ADD, UPDATE, MERGE, APPEND, REPLACE, REWRITE, and SKIP should be appropriate given the relevant existing memories. Penalize duplicate ADDs, unnecessary updates, overwriting useful facts, and failing to update changed facts.
4. Module fit: judge whether the output belongs in this memory module.
   - core: stable user identity, relationships, preferences, values, long-term goals, recurring traits.
   - episodic: timestamped events and experiences.
   - semantic: stable facts about people, places, objects, and concepts in the user's world.
   - procedural: reusable procedures, workflows, recipes, or instructions. If there is no procedural content, SKIP or an empty operation can be best.
5. Specificity and retrievability: concise but concrete memories with good anchors are better than vague summaries.
6. Non-redundancy: avoid duplicate, bloated, or low-value memories.

Important rules:
- The gold answers are hints about important facts, not permission to add facts unsupported by this session.
- First identify which QA targets are actually supported by this conversation session; ignore unrelated targets when choosing a winner.
- In each module output, existing_memory_count and relevant_existing_memories are context only; judge the current operations and newly written/updated memory content.
- Do not prefer longer output or more operations by default.
- Do not reward style unless it improves retrieval or factual clarity.
- If this session is unrelated to the listed QA targets, judge long-term memory usefulness for future QA.
- If both are similarly good or similarly bad, choose Tie.
- Use the scoring fields to make the comparison explicit, but the final winner should follow the priority order above.

Conversation session:
{session}

Most relevant downstream QA targets:
{questions}

Gold answer(s):
{gold_answers}

Memory module scope:
{module_scope}

System A output:
{system_a_steps}

System B output:
{system_b_steps}

Return JSON only:
{{
  "scores": {{
    "evidence_grounded_faithfulness": {{"A": 1, "B": 1}},
    "answer_critical_coverage": {{"A": 1, "B": 1}},
    "operation_rationality": {{"A": 1, "B": 1}},
    "module_fit": {{"A": 1, "B": 1}},
    "retrievability": {{"A": 1, "B": 1}},
    "non_redundancy": {{"A": 1, "B": 1}}
  }},
  "winner": "A" | "B" | "Tie",
  "reason": "one concise sentence"
}}
\end{lstlisting}
\end{tcolorbox}
\caption{Prompt template for evaluating the quality of intermediate memory-construction outputs in a long-term memory QA system.}
\label{fig:memory_construction_evaluation_prompt}
\end{figure*}

\mypar{Hyperparameters.}
Table~\ref{tab:hparams} lists all training hyperparameters.

\begin{table}[!t]
  \centering
  \small
  \begin{tabular}{ll}
    \toprule
    Hyperparameter & Value \\
    \midrule
    Backbone & Qwen3-4B \\
    SFT learning rate & 5e-7 \\
    RL learning rate & 1e-6 \\
    SFT batch size & 32 \\
    RL effective batch size & 256 \\
    GRPO group size & 8 \\
    Training steps (SFT) & 3000 \\
    Training steps (RL) & 400 \\
    Max sequence length & 6000 \\
    \bottomrule
  \end{tabular}
    \caption{Training hyperparameters for \ourshort{}.}
  \label{tab:hparams}
\end{table}

\mypar{Compute.}
We use 8$\times$ NVIDIA H800 (80GB) GPUs.
For \ourshort{}, SFT training time is $\sim$30 hours while RL training time is $\sim$5 days.

\mypar{Reproducibility.}
Code and trained model checkpoints will be released upon acceptance.
The LongMemEval, LoCoMo, and PerLTQA datasets are all
publicly available; download instructions are included in the repository.

\section{Extended Related Work}
\mypar{Process rewards for agents via attribution.}
Reward sparsity is a significant challenge in long-horizon reasoning and agentic tasks because outcome rewards are easier to obtain than process rewards.
Process rewards can reduce redundancy, improve problem-solving efficiency, and reinforce critical intermediate milestones.
Among several sources of process supervision, attribution is a prominent direction that estimates the contribution of intermediate steps to final outcomes.
Tree-branch approaches, exemplified by \textit{Agentic Reinforced Policy Optimization (ARPO)}~\citep{dong2026agentic}, reward intermediate steps according to the likelihood that their downstream branches lead to correct outcomes.
HCAPO~\citep{tan2026hindsight} instead uses an LLM as a post-hoc critic to refine step-level Q-values through hindsight reasoning and combines these estimates with multi-scale advantages for long-horizon agent RL.
For multi-agent prompt optimization, \citet{li2026unifying} decompose credit along temporal and structural axes, using critic-generated proxy gradients to identify weak rounds and roles and verbalized block coordinate descent to update the corresponding prompts.
These approaches assign credit at the step, round, or role level to guide policy or prompt updates, whereas \ourshort{} derives answer-conditioned, counterfactual token-level rewards for memory-action tokens.

Our work is among the first to derive process rewards from context attribution.
One particularly relevant concurrent work, \textit{Leave-One-Turn Attribution for Policy Optimization (LOTAPO)}~\citep{zhu2026lapo}, estimates the contribution of each turn to the final gold answer.
Whereas LOTAPO derives \textit{action-level} process rewards from attribution, \ourshort{} provides finer-grained \textit{token-level} rewards.
We further show in \cref{tab:reward_granularity_ablation} that token-level rewards outperform action-level rewards.

\mypar{Failure attribution for agents.}
\citet{zhang2025which} formalize automated failure attribution for LLM multi-agent systems and introduce the Who\&When benchmark for identifying the agent and step responsible for a failed trajectory.
\citet{yeh2026tracing} propose OAT, which uses neural controlled differential equations to learn the latent dynamics of successful trajectories and identify anomalous steps in failed trajectories without step-level supervision on failure data.
These methods diagnose failures by localizing responsible steps, whereas \ourshort{} uses final-answer contribution during training to assign token-level rewards to memory-action tokens.

\mypar{Attribution to reasoning steps.}
\citet{pan2026towards} propose recursive attribution for faithful and efficient attribution over long reasoning traces in reasoning LLMs.

\end{document}